\documentclass[letterpaper, 10pt, conference]{ieeeconf}  

\usepackage{times}
\usepackage{epsfig}
\usepackage{graphicx}
\usepackage{amsmath}
\usepackage{amssymb}
\usepackage[pagebackref=true,breaklinks=true,letterpaper=true,colorlinks,bookmarks=false]{hyperref}
\usepackage{mathptmx}
\usepackage{amsopn}
\usepackage{algorithm}
\usepackage{algorithmic}
\usepackage{multirow}
\usepackage{epstopdf}
\usepackage[caption=false]{subfig}
\usepackage[flushleft]{threeparttable}

\IEEEoverridecommandlockouts                              
\title{\LARGE \bf
eBear: An Expressive Bear-Like Robot
}

\author{Xiao Zhang, Ali Mollahosseini, Amir H. Kargar B., Evan Boucher, \\ Richard M. Voyles, Rodney Nielsen and Mohammd H. Mahoor
\thanks{*This work is supported by the NSF grant IIS-1111568.}
\thanks{Xiao Zhang, Ali Mollahosseini, Amir H. Kargar B., Evan Boucher, and Mohammd H. Mahoor are with the Department of Electrical and Computer Engineering at University of Denver, Denver, CO 80210. E-mails: {\tt\small \{xzhang62, ali.mollahosseini, akargarb, evan.boucher, mmahoor\}@du.edu}}
\thanks{Richard M. Voyles is with the College of Technology at Purdue University, West Lafayette, IN 47907. E-mail: {\tt\small rvoyles@purdue.edu}}
\thanks{Rodney Nielsen is with the Department of Computer Science and Engineering at University of North Texas, Denton, TX 76203. E-mail: {\tt\small rodney.nielsen@unt.edu}}
}

\begin{document}

\maketitle
\thispagestyle{empty}
\pagestyle{empty}

\begin{abstract}

This paper presents an anthropomorphic robotic bear for the exploration of human-robot interaction including verbal and non-verbal communications. This robot is implemented with a hybrid face composed of a mechanical faceplate with 10 DOFs and an LCD-display-equipped mouth. The facial emotions of the bear are designed based on the description of the Facial Action Coding System as well as some animal-like gestures described by Darwin. The mouth movements are realized by synthesizing emotions with speech. User acceptance investigations have been conducted to evaluate the likability of these facial behaviors exhibited by the eBear. Multiple Kernel Learning is proposed to fuse different features for recognizing user's facial expressions. Our experimental results show that the developed Bear-Like robot can perceive basic facial expressions and provide emotive conveyance towards human beings.

\end{abstract}

\section{Introduction}
\label{intro}
It is an exciting time in robotics. Different types of robots, ranging from industrial robots to human-like androids with a variety of functionalities, have been developed. Although robots are finding their place as artificial pets (e.g. Leonardo~\cite{leonardo_skin}), entertainers (e.g. NAO from Aledebaran Robotics), and tools for therapists (e.g. PARO~\cite{paro}), current technologies have yet to reach the full emotional and social capabilities necessary for rich and robust interaction with human beings. There is a common agreement among neuroscientists and psychologists that emotions are necessary for robust decision making processes in humans. There is also a strong belief among robotics researchers that ``emotions are needed for robots to be social'' and ``to effectively communicate with humans''~\cite{puaics2010emotions}. Producing expressive robots that connect to people at a primal level will lead not only to more productive machines, but also to a deeper understanding of the human condition and intelligence.

This paper presents our current progress at the University of Denver robotics lab in developing a bear-like robotic head called expressive Bear (eBear). The eBear can show animal-like facial expressions, ear movement, eye gaze attention, and head gestures. It can understand user's facial expressions through a camera on the head with our proposed recognition algorithm. Our robot can react to perceived users' emotions via expression mirroring. The eBear has also the ability to speak with accurate lip movements. The eBear's face is a hybrid of mechanical actuators and computer graphics. In particular, we use computer graphics to design facial expressions appearing on the mouth and accurate visual speech during spoken dialog. The animal-like expressions are designed using Darwin's interpretation of the emotive expressions in animals~\cite{darwin1998expression} as well as the Facial Action Coding System (FACS) developed by Paul Ekman~\cite{ekman:friesen:1978}.

The remainder of this paper is organized as follows: Section~\ref{related} reviews the related work. Section~\ref{hardware} describes the hardware design and the facial expression generation of eBear. Section~\ref{method} presents our proposed method for human expression recognition. Section~\ref{expt} shows and discusses the experimental results for evaluating the performance of our robot platform. Finally, Section~\ref{concld} concludes the paper.

\section{Related work}
\label{related}
We review the following well-developed expressive robotic systems. Breazeal's research group was one of the first to develop an expressive anthropomorphic robot ``Kismet"~\cite{breazeal2003emotion}, which engages people in natural and affective face-to-face communication. Equipped with 16 degrees of freedom (DOFs), Kismet can show a wide assortment of facial expressions, which reflect its emotional state. Later, the robot ``Leonardo" was developed with 32 DOF's on the face and is capable of behaving near-human facial expressions. One of the crucial design observations taken from these robots is that ear movements play an important part in emotional and nonverbal expression interaction~\cite{breazeal2004tutelage, hri2002}. To this end, the robot ``Meka'' was upgraded to ``Simon'' by adding ear movements on its head~\cite{diana2011shape}. In addition, another animal-like robot ``iCat'' \cite{iCat} was developed by Philips as an experimentation platform for human-robot interaction research. The iCat can generate multiple facial expressions to conduct social affective communication with human users.

Although much progress has been made to design aesthetic animal-like faces of these articulated robots, it is usually difficult to design the mechanic structures with multiple DOFs and jointly control them when displaying several complex facial behaviors such as mouth movement. In this case, the robot ``BERT2"~\cite{bazo2010design} was designed with a hybrid face including a plastic faceplate and an LCD display, where eye brows, eyes and mouth movement were displayed via graphics animation. The recognizability of BERT2's facial expressions were evaluated and verified through human-robot interaction experiments. Compared to BERT2, our eBear has 10 DOFs on the mechanical faceplate including ears, and it is able to blend the expressive movement with speech during mouth animation, which incorporates the benefits of both mechanical and graphical facial elements.

\section{eBear design and implementation}
\label{hardware}
In this section, we introduce eBear's hardware system including the design of mechanical features, the mouth animation methods on an LCD display and the methodology of generating robot's facial expressions.

\subsection{Mechanical design}
\begin{figure}[thpb]
  \centering
  \includegraphics[scale=0.35]{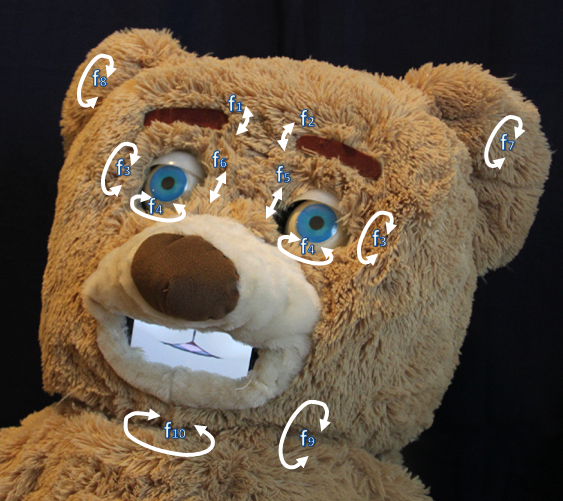}\\ 
  \caption{10 DOFs of eBear's mechanical faceplate}
  \label{fig:head}
\end{figure}

\begin{figure}[h]
\centering
\subfloat[]
{
	\centering
    \includegraphics[scale = 0.60]{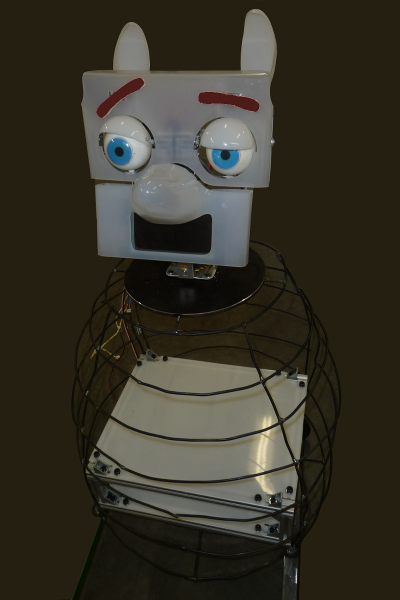}
}
\subfloat[]
{
    \includegraphics[scale = 0.36]{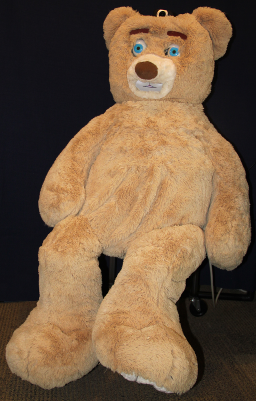}
}
\caption{\label{ebear}
Mechanic platform of eBear
}
\end{figure}

The eBear's head consists of four elements on the mechanical faceplate: eyebrows, eyeballs, eyelids and ears, with a total of 10 DOFs. As shown in Fig. \ref{fig:head}, the DOFs are: left and right eyebrows' angle (row) $f_{1}$ and $f_{2}$; forehead tilt $f_{3}$ and eye balls yaw $f_{4}$; left and right eyebrow eyelids' openness/closeness (pitch) $f_{5}$ and $f_{6}$; left and right ears pitch $f_{7}$ and $f_{8}$; neck pitch $f_{9}$ and yaw $f_{10}$.
Each DOF is controlled via one ``Hitec" PWM servomotor that is properly attached to the head system. The servomotors receive commands from a ``Mini Maestro Servomotor Controller" unit programmed using C\# .Net Framework, which can be controlled cooperatively to show multiple desired facial expressions with different dynamics (angle of rotations, speed and accelerations). Fig. \ref{ebear} shows our developed robotic head mounted on a pedestal box covered by bear-like fur.

\subsection{Mouth design}
\label{mouth_animation}
Mechanical lips usually lack enough DOFs to show expressions and synchronizing lip movements with speech, especially when the robot speaks fast. To this end, we used an LCD display (4.3$''$ TFT LCD panel by Sharp) to animate lip movements. The LCD has $480 \times 272$ pixel resolution in 16 bit colors (RGB565) and can be programmed by OpenGL via provided APIs at 80-90 full frames per second over USB. 

To create an accurate natural visual speech and show expression, we developed a mouth animation software based on multi-target morphing method \cite{ma2004animating}. For a given language, visually similar phonemes are grouped into units called visemes. We categorized English phonemes into 20 viseme classes. For example the constants /b/, /p/ and /m/ in the words ``buy,'' ``pie,'' and ``my'' form a single viseme class. These classes represent the articulation targets that lips and tongue move to during speech production. A graphic artist designed 3D models of these visemes in Maya and a natural visual speech is obtained by blending proper models corresponded to each part of speech with different weights.

Recorded utterances are processed by the Bavieca speech recognizer \cite{bolanos2012bavieca}, which receives the sequence of words and the speech waveform as input, and provides a time-aligned phonetic transcription of the spoken utterance. The aligned phonemes are represented using the International Phonetic Alphabet (IPA), a standard that is used to provide a unique symbolic notational for the realization of phonemes in all of the world's languages \cite{international1999handbook}. As IPA is intended as a standard for the phonemic and phonetic representation of all spoken languages, having IPA in our system will allow us to add other languages easily as long as the speech recognizer is trained for that language. During speech production, the animation system receives the time-aligned phonetic input from Bavieca system, converts the phonetic symbols into the corresponding visemes, which specifies the movements of the mouth and tongue, synchronized with the recorded or synthesized speech.

To achieve a smooth and realistic look, coarticulation are modeled by smoothing across adjacent phonemes, while adjusting the kernel to assure that certain phonemes (e.g., /b/, /p/ and /m/) always reach their target. The kernel smoothing technique~\cite{eubank1999nonparametric} is one of the most common nonparametric techniques to estimate probability density and smooth data in statistics. We used the Epanechnikov kernel~\cite{epanechnikov1969non} to pull the weights for each viseme associated with the current time value and set the weights for those visemes' morph targets. Using the kernel technique resulted in smoother and more natural looking animations; however, when utterances included the labial phonemes /b/, /m/, /p/, which are accompanied by lip closure, the smoothing algorithm prevented the lips from closing when adjacent phoneme targets caused the lips to be open (e.g., /a/ as in ``mama''). To force lip closure for the labials, we extended the duration of labial visemes for /b/ and /p/ to include the closure interval, thus increasing the chance that at least one frame consisting of just the labial viseme will appear.

\begin{figure*}
\centering
\subfloat[][\label{Anger}Anger]
{
	\centering
    \includegraphics[scale = 0.12]{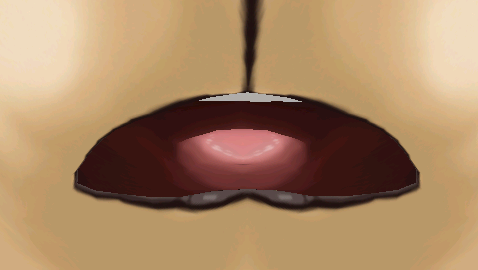}
}
\subfloat[][\label{Joy} Joy]
{
	\centering
    \includegraphics[scale = 0.12]{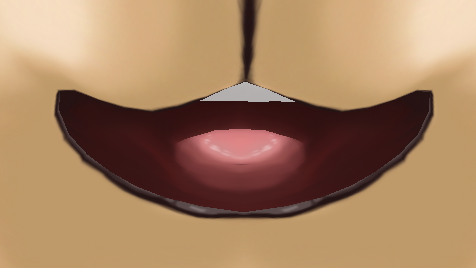}
}
\subfloat[][\label{Sad} Sadness]
{
	\centering
    \includegraphics[scale = 0.12]{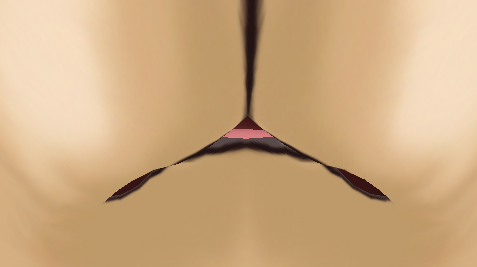}
}
\subfloat[][\label{Disgust} Disgust]
{
	\centering
    \includegraphics[scale = 0.12]{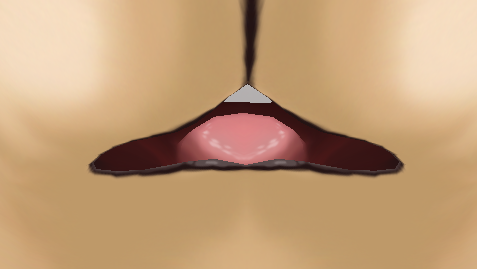}
}
\subfloat[][\label{Surprise}Surprise]
{
	\centering
    \includegraphics[scale = 0.12]{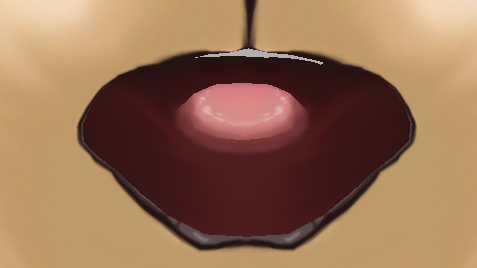}
}
\subfloat[][\label{Fear} Fear]
{
	\centering
    \includegraphics[scale = 0.12]{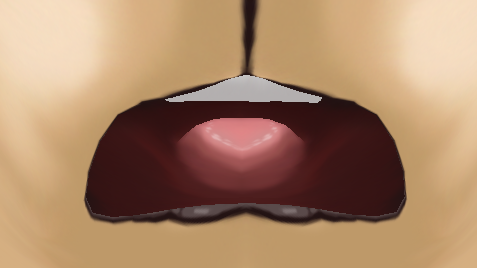}
}
\subfloat[][\label{Neutrality}Neutral]
{
	\centering
    \includegraphics[scale = 0.12]{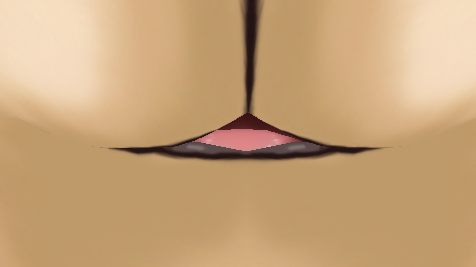}
}
\subfloat[][\label{Viseme o}Viseme `O']
{
	\centering
    \includegraphics[scale = 0.12]{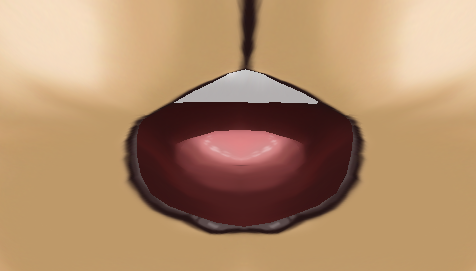}
}
\subfloat[][\label{Visime m} Visime `m']
{
	\centering
    \includegraphics[scale = 0.12]{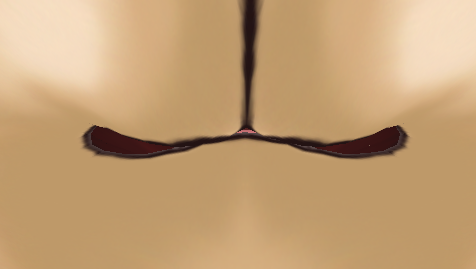}
}
\caption{\label{fig:SomeVisemesExpression}
Examples of some visemes and expressions
}
\end{figure*}

In order to synthesize expressions with lip movements, we blend the current viseme with the desired expression as:
\begin{equation}
\label{mouth}
F_{j}=F_{c}+ \lambda_j (F_j^{max} - F_0)
\end{equation}
where $F_{c}$ represents the current viseme, $F_j^{max}$ is the desired expression model at the maximum intensity, $F_0$ is the Neutral model. The parameter $\lambda_j \in [0,1]$ is the intensity of the $j^{th}$ expression model $F_{j}$. We designed 3D mouth models of the six basic expressions at their maximum intensity in Maya based on FACS. In the FACS system each facial expression is defined as a combination of several facial Action Units (AUs). For example Joy involves lip corner puller (AU12) and Sadness involves lip corner depressor (AU15). Fig. \ref{fig:SomeVisemesExpression} demonstrates these basic expressions and some visemes used in our animation system.

\subsection{eBear's facial expression generation}
eBear's facial expressions are designed based on the description of FACS and animal-like gestures. According to the methodology described in \cite{user_acceptance}, all the AUs should be projected into DOFs available on the robot's face. In Tab. \ref{tab:DOF_Expression}, the first two columns show six basic emotions as well as the related AUs. The third column illustrates the DOFs that are used for each emotion. AUs that can be displayed on the robot are indicated in bold. Disgust is the most difficult emotion to express in animal-like robots, since nose wrinkle in Disgust is tough to be embodied in robots. In eBear $f_{3}$ (forehead tilt) is used to show such wrinkles. We call the expressions that are only based on the FACS ``AU based".

\begin{figure*}
\centering
\subfloat[][\label{joy_au}Joy]
{
    \includegraphics[scale = 0.18]{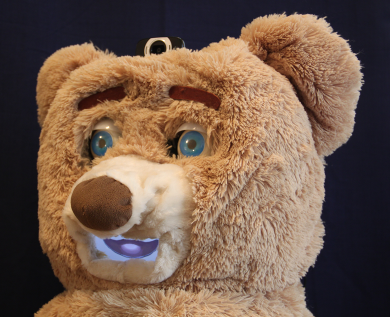}
}
\subfloat[][\label{sad_au}Sadness]
{
    \includegraphics[scale = 0.18]{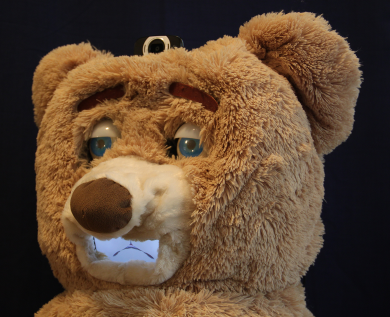}
}
\subfloat[][\label{fear_au}Fear]
{
    \includegraphics[scale = 0.18]{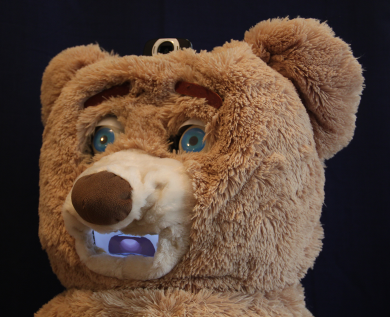}
}
\subfloat[][\label{disgust_au}Disgust]
{
    \includegraphics[scale = 0.18]{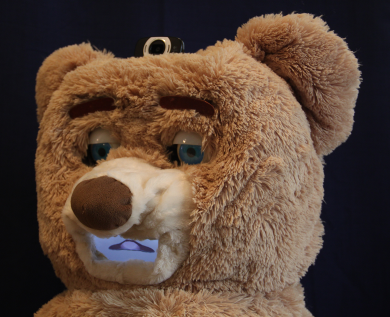}
}
\subfloat[][\label{anger_au}Anger]
{
    \includegraphics[scale = 0.18]{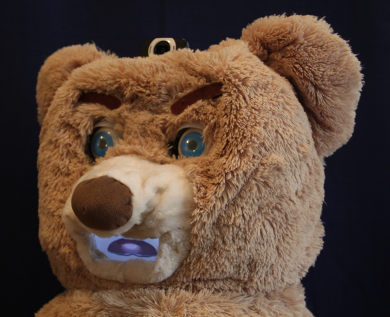}
}
\subfloat[][\label{surprise_au}Surprise]
{
    \includegraphics[scale = 0.18]{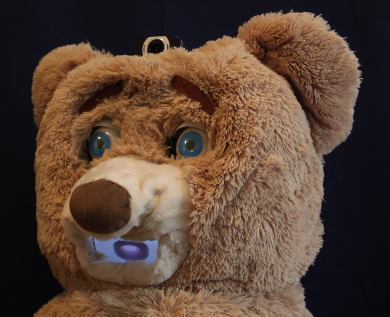}
}
\\

\subfloat[][\label{joy_dyn} Joy]
{
    \includegraphics[scale = 0.18]{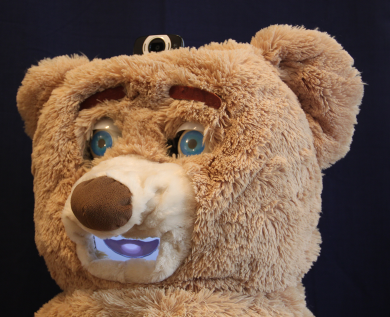}
}
\subfloat[][\label{sad_dyn}Sadness]
{
	\includegraphics[scale = 0.18]{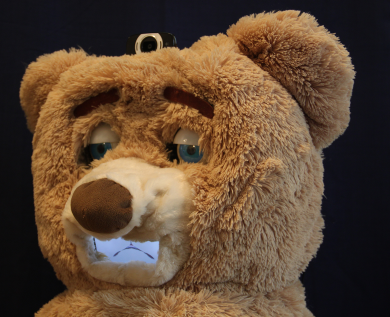}
}
\subfloat[][\label{fear_dyn}Fear]
{
    \includegraphics[scale = 0.18]{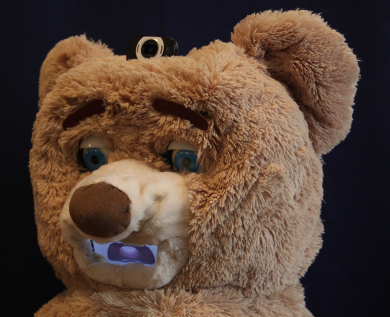}
}
\subfloat[][\label{disgust_dyn}Disgust]
{
    \includegraphics[scale = 0.18]{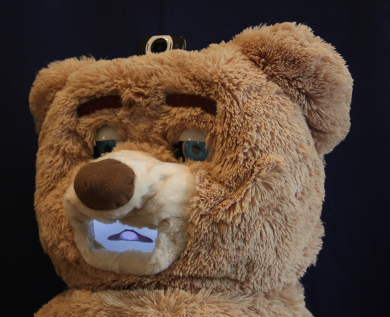}
}
\subfloat[][\label{anger_dyn}Anger]
{
    \includegraphics[scale = 0.18]{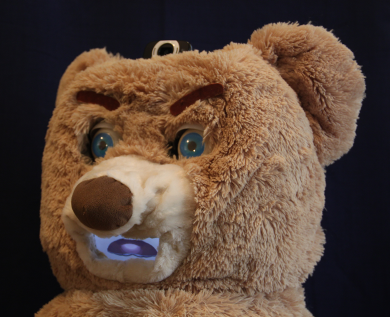}
}
\subfloat[][\label{surprise_dyn}Surprise]
{
    \includegraphics[scale = 0.18]{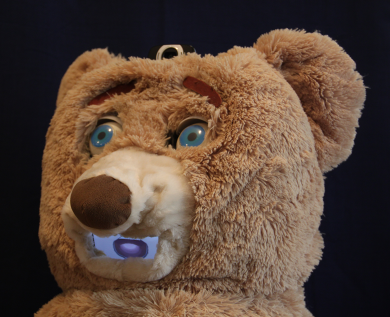}
}

\caption{\label{fig:ExpressionOnEbear}
Examples of expressions on eBear (Row 1: AU based; Row 2: AU+Animal based).
}
\end{figure*}

Besides the AU based expressions, we were inspired by animals' way of expressing emotions to improve eBear's expressivity. Darwin \cite{darwin1998expression} studied and described emotions in terms of various body organs, and concluded that ear movements are highly expressive in many animals. For instance, dogs have different ear movements in particular situations. When they are pleased or being caressed, the ears are usually drawn back and fall down slightly. In Fear, ears are also drawn back, but not pressed closely to the head. The other common situation among many animals is to draw back their ears when they feel savage. The above facts as well as ideas from cartoon animations have been taken into consideration in designing animal-like facial expression generations in the eBear. For instance, in happiness, continuous movements of ears in reverse directions (i.e. each ear moves independently between 0.5 to 1.5 seconds depending on the expression intensity) have been used to show the Joy of the bear (Fig. \ref{fig:ExpressionOnEbear}\subref{joy_dyn}). In Disgust, ears are drawn forward the combination of ears and forehead tilt makes nose wrinkles more visible (Fig. \ref{fig:ExpressionOnEbear}\subref{disgust_dyn}). In Anger, ears become erected to display the Anger of the bear (Fig. \ref{fig:ExpressionOnEbear}\subref{anger_dyn}). Be drawn forward in Sadness (Fig. \ref{fig:ExpressionOnEbear}\subref{sad_dyn}) and going back in Surprise (Fig. \ref{fig:ExpressionOnEbear}\subref{surprise_dyn}) are some other situations, where ear movements have been used. In Fear, same as dogs, ears are also drawn back and forehead tilt is used to looking down (Fig. \ref{fig:ExpressionOnEbear}\subref{fear_dyn}). We call this modified version which is a combination of AUs, ear movement and forehead tilt as ``AU+Animal based" in the rest of this paper. 
\begin{table}[htpb]
\caption{DOFs in AU based and AU+Animal based expressions}
\vspace{-0.3cm}
\label{tab:DOF_Expression}
\centering
\begin{tabular}{p{1cm}|p{2.2cm}|p{1.3cm}|p{2.4cm}|}
\cline{2-4}
\multicolumn{1}{c}{\multirow{2}{*}{\textbf{}}} & \multicolumn{1}{|c}{\multirow{2}{*}{\scriptsize{\textbf{FACS AUs}}}} & \multicolumn{2}{|c|}{\scriptsize{\textbf{Corresponding DOF}}}                                        \\ \cline{3-4} 
                                               &                                                         & \multicolumn{1}{|c}{\scriptsize{\textbf{AU based}}} & \multicolumn{1}{|c|}{\scriptsize{\textbf{AU+Animal based}}} \\ \hline
\multicolumn{1}{|l|}{\scriptsize{Happiness}} & \scriptsize{AU6, \textbf{AU12}}          & \scriptsize{L}\textsuperscript{*}         & \scriptsize{L, $f_7$, $f_8$}             \\ \hline
\multicolumn{1}{|l|}{\scriptsize{Sadness}}   & \scriptsize{\textbf{AU1}, AU4, \textbf{AU15}}      & \scriptsize{L, $f_1$, $f_2$, $f_5$, $f_6$} & \scriptsize{L, $f_1$, $f_2$, $f_3$, $f_5$, $f_6$, $f_7$, $f_8$}   \\ \hline
\multicolumn{1}{|l|}{\scriptsize{Fear}}      & \scriptsize{\textbf{AU1}, \textbf{AU2}, AU4, \textbf{AU5}, \textbf{AU20}, \textbf{AU26}} &\scriptsize{L, $f_1$, $f_2$, $f_5$, $f_6$} & \scriptsize{L, $f_1$, $f_2$, $f_3$, $f_5$, $f_6$, $f_7$, $f_8$}   \\ \hline
\multicolumn{1}{|l|}{\scriptsize{Disgust}}   & \scriptsize{\textbf{AU9}, \textbf{AU15}, \textbf{AU16}}       &\scriptsize{L, $f_3$}       & \scriptsize{L, $f_3$, $f_7$, $f_8$}           \\ \hline
\multicolumn{1}{|l|}{\scriptsize{Anger}}     & \scriptsize{AU4, \textbf{AU5}, AU7, \textbf{AU23}}      & \scriptsize{L, $f_1$, $f_2$, $f_5$, $f_6$} & \scriptsize{L, $f_1$, $f_2$, $f_5$, $f_6$, $f_7$, $f_8$}     \\ \hline
\multicolumn{1}{|l|}{\scriptsize{Surprise}}  & \scriptsize{\textbf{AU1}, \textbf{AU2}, \textbf{AU5B}, \textbf{AU26}}     & \scriptsize{L, $f_1$, $f_2$, $f_5$, $f_6$} & \scriptsize{L, $f_1$, $f_2$, $f_3$, $f_5$, $f_6$, $f_7$, $f_8$, $f_9$} \\ \hline
\end{tabular}
    \begin{tablenotes}
      \item \textsuperscript{*} \scriptsize{L stands for corresponding AUs that are shown on LCD}
    \end{tablenotes}
\end{table}

The eBear is designed to show facial expressions with multiple intensities. For each DOF $f_{i},(i=1,2,...,10)$ involved in a desired expression $e_{j}$, we denote $f_{i}^{0}$ as the value corresponding to its neutral state and $f_{i,j}^{max}$ for its expressive state at the maximum intensity. Then the movement of DOF can be formulated as:
\begin{equation}
\label{head}
f_{i,j}=f_{i}^{0}+ \mu_{i,j}(f_{i,j}^{max}-f_{i}^{0})
\end{equation}
where, $\mu_{i,j} \in [0,1]$ represents the intensity of $f_{i}$ in expression $e_{j}$. Hence, together with the mouth animation described in Eq. \ref{mouth}, we assign the value of $\mu_{i,j}$ and $\lambda_{j}$ to jointly control the mouth and mechanical facial elements for expressing the robot's facial emotions with various intensities.

\section{Facial expression perception}
\label{method}
Facial expression recognition plays an important part in robotic systems. Accurate recognition results can help social robots to achieve right understanding of users' emotive states and perform proper reactions. As different facial features have different distributions, especially in multiclass classification tasks, using one type of features may not be distinguishable for all classes. In eBear, we utilize our proposed HessianMKL based multiclass-SVM (Support Vector Machines) \cite{Xiao_FG} to recognize six basic human expressions and neutral faces. The idea of our method is to fuse different types of facial features with multiple kernels and increase the discriminative power of canonical multiclass-SVM. 

Given $N$ training samples, $x_{i} \in R^{D},(i=1,2,...,N)$ is the $i^{th}$ feature vector. $\{k^{m}(\cdot,\cdot)\}_{m=1}^{M}$ indicates the utilized $M$ basis kernels. Suppose we classify $P$ classes with one-against-one strategy, then $P(P-1)/2$ binary classifiers are built for all pairwise classes. $\Phi$ denotes the set of all pairs of distinct classes in the multiclass problem, and $y_{i}^{(p)} \in \{-1,+1\}, (p \in \Phi)$ is the label of $i^{th}$ sample in $p^{th}$ binary classification problem. Our proposed method can be formulated as:
\begin{equation}
\label{dual_hessianmkl}
\begin{split}
& \forall p \in \Phi, \hspace{1mm} \mathop{\min}_{d_{m}^{(p)}\geq 0} \hspace{1mm} \mathop{\max}_{0 \leq \alpha_{i}^{(p)}\leq C} \hspace{2mm} J(d^{(p)},\alpha^{(p)}) = \\
& \hspace{10mm} \sum_{i=1}^{N}\alpha_{i}^{(p)} - \hspace{.5mm} \frac{1}{2} \sum_{i,j=1}^{N}\alpha_{i}^{(p)}\alpha_{j}^{(p)}y_{i}^{(p)}y_{j}^{(p)}(\sum_{m=1}^{M}d_{m}^{(p)}K^{m})\\
& \hspace{3mm} s.t. \hspace{3mm} \sum_{i=1}^{M}d_{m}^{(p)} = 1, \hspace{1mm} \sum_{i=1}^{N}\alpha_{i}^{(p)}y_{i}^{(p)}=0 \\
\end{split}
\end{equation}
where $\alpha^{(p)} = (\alpha_{1}^{(p)}, \alpha_{2}^{(p)}, \cdots, \alpha_{N}^{(p)})^{T}$ is the vector of Lagrangian dual variables corresponding to each training sample, and $C$ is a positive constant preset to control the relative influence of nonseparable samples in SVM. $d^{(p)} = (d_{1}^{(p)},d_{2}^{(p)},\cdots,d_{M}^{(p)})^{T}$ is the kernel combination weight vector, and $K_{i,j}^{m}=k^{m}(x_{i},x_{j})$. The Algorithm 1 in \cite{Xiao_FG} shows the solving procedure. Once $\alpha^{(p)}$ and $d^{(p)}$ are determined, the learned discriminant hyperplane of each binary classifier is
\begin{equation}
\label{mkl_hyperplane}
h^{(p)}(x)=\sum_{m=1}^{M}d_{m}^{(p)}(\sum_{i=1}^{N}{{\alpha_{i}^{(p)}}y_{i}^{(p)}k^{m}(x_{i},x)}), \hspace{1mm} \forall p \in \Phi
\end{equation}
where $x \in R^{D}$ is a given test sample. The class label of $x$ is assign by a max-wins voting strategy across all binary classifiers in the multiclass classification problem. 

\section{Experiments}
\label{expt}
This section illustrates the details of three experiments on eBear. Based on the user acceptance investigation, we verify the validity of our robot expression and visual speech design. Our proposed expression recognition method is evaluated on the CK+ \cite{lucey2010extended} and the MMI face databases \cite{ValstarPantic2010}. The experiment of user expression imitation is set to evaluate the performance of our platform in dynamic environments.

\subsection{Human perception of eBear's facial expressions}
21 subjects between ages of 21 to 51 years participated in our user acceptance experiment. They were from various races and cultural backgrounds and had never been exposed to the eBear before the experiment. The subjects were sitting in front of the eBear at the distance of 1 meter. 12 facial expressions were presented to each subject in a random order. Six of the expressions were AU based and the other six ones were AU+animal based. The robot started with the neutral expression and then switch to one of the six basic emotions (Joy, Anger, Sadness, Disgust, Surprise, Fear). Each expression took around 1.5 seconds to reach its final position. Participants could take as much time as they required to recognize the expressions. They were then asked to select one of the six basic emotions or Neutral available on the questionnaire. They could also respond ``none," if they were unable to assign the facial expression to any category.

The confusion matrices of AU based and AU+animal based expressions are shown in Tab. \ref{nDynamic_cMatrix} and Tab. \ref{Dynamic_cMatrix}, respectively. The recognition rates of Joy, Anger, Disgust and Surprise were increased in the AU+animal based mode from the AU based mode while the recognition rate of Sadness was decreased by around 4.8\%. This is because the participants confused it with Disgust. One possible reason for this is that, in AU based mode Sadness is shown as AU1 + AU15, but in AU+animal based mode the forehead of the eBear was also tilted downward which is a common action associated with Disgust expression. Fear was recognized 33.3\% of the times in AU based mode and was mainly confused with Surprise. In AU+animal based mode Fear was recognized 19.0\% of the times and was often confused with Sadness, and lowering the forehead in Fear and Sadness may be the reason for this confusion. Participants may have perceived Fear as an alternative form of Sadness, although the position of eyebrows are completely different.

\begin{table}[htpb]
\caption{AU based expressions confusion matrix}
\vspace{-0.3cm}
\label{nDynamic_cMatrix}
\centering
\begin{tabular}{|p{0.8cm}|p{0.5cm}|p{0.5cm}|p{0.5cm}|p{0.5cm}|p{0.5cm}|p{0.5cm}|p{0.5cm}|p{0.5cm}|}
\hline
\scriptsize{\%}       & \scriptsize{Jy}   & \scriptsize{Ag} & \scriptsize{Sd}   & \scriptsize{Dg} & \scriptsize{Sp} & \scriptsize{Fr}  & \scriptsize{Nt} & \scriptsize{Nn} \\ \hline
\scriptsize{Joy}      & \scriptsize{\textbf{90.5}}  & \scriptsize{0}     & \scriptsize{0}     & \scriptsize{0}       & \scriptsize{9.5}      & \scriptsize{0}     & \scriptsize{0}       & \scriptsize{0}    \\ \hline
\scriptsize{Anger}    & \scriptsize{0}     & \scriptsize{\textbf{80.9}}  & \scriptsize{0}     & \scriptsize{4.8}     & \scriptsize{14.3}     & \scriptsize{0}     & \scriptsize{0}       & \scriptsize{0}    \\ \hline
\scriptsize{Sadness}      & \scriptsize{0}     & \scriptsize{0}     & \scriptsize{\textbf{100}}   & \scriptsize{0}       & \scriptsize{0}        & \scriptsize{0}     & \scriptsize{0}       & \scriptsize{0}    \\ \hline
\scriptsize{Disgust}  & \scriptsize{0}     & \scriptsize{0}     & \scriptsize{80.9}  & \scriptsize{\textbf{9.5}}     & \scriptsize{0}        & \scriptsize{0}     & \scriptsize{4.8}     & \scriptsize{4.8}  \\ \hline
\scriptsize{Surprise} & \scriptsize{0}     & \scriptsize{0}     & \scriptsize{14.3}  & \scriptsize{0}       & \scriptsize{\textbf{61.9}}     & \scriptsize{23.8}  & \scriptsize{0}       & \scriptsize{0}    \\ \hline
\scriptsize{Fear}     & \scriptsize{4.8}   & \scriptsize{0}     & \scriptsize{14.3}  & \scriptsize{19.0}    & \scriptsize{28.6}     & \scriptsize{\textbf{33.3}}  & \scriptsize{0}       & \scriptsize{0}    \\ \hline
\end{tabular}
    \begin{tablenotes}
      \item \textsuperscript{*} \scriptsize{Nt and Nn stand for Neutral and None respectively}
    \end{tablenotes}
\end{table}
\vspace{-0.5cm}
\begin{table}[htpb]
\caption{AU+animal based Expressions confusion matrix}
\vspace{-0.3cm}
\label{Dynamic_cMatrix}
\centering
\begin{tabular}{|p{0.8cm}|p{0.5cm}|p{0.5cm}|p{0.5cm}|p{0.5cm}|p{0.5cm}|p{0.5cm}|p{0.5cm}|p{0.5cm}|}
\hline
\scriptsize{\%}       & \scriptsize{Jy}   & \scriptsize{Ag} & \scriptsize{Sd}   & \scriptsize{Dg} & \scriptsize{Sp} & \scriptsize{Fr}  & \scriptsize{Nt} & \scriptsize{Nn}  \\ \hline
\scriptsize{Joy}      & \scriptsize{\textbf{95.2}}  & \scriptsize{0}     & \scriptsize{0}     & \scriptsize{0}       & \scriptsize{0}        & \scriptsize{0}     & \scriptsize{4.8}     & \scriptsize{0}     \\ \hline
\scriptsize{Anger}    & \scriptsize{0}     & \scriptsize{\textbf{85.7}}  & \scriptsize{0}     & \scriptsize{0}       & \scriptsize{4.8}      & \scriptsize{9.5}   & \scriptsize{0}       & \scriptsize{0}     \\ \hline
\scriptsize{Sadness}  & \scriptsize{0}     & \scriptsize{0}     & \scriptsize{\textbf{95.2}}  & \scriptsize{4.8}     & \scriptsize{0}        & \scriptsize{0}     & \scriptsize{0}       & \scriptsize{0}     \\ \hline
\scriptsize{Disgust}  & \scriptsize{0}     & \scriptsize{0}     & \scriptsize{23.8}  & \scriptsize{\textbf{42.8}}    & \scriptsize{4.8}      & \scriptsize{9.5}   & \scriptsize{4.8}     & \scriptsize{14.3}  \\ \hline
\scriptsize{Surprise} & \scriptsize{0}     & \scriptsize{0}     & \scriptsize{0}     & \scriptsize{0}       & \scriptsize{\textbf{71.4}}     & \scriptsize{28.6}  & \scriptsize{0}       & \scriptsize{0}     \\ \hline
\scriptsize{Fear}     & \scriptsize{0}     & \scriptsize{0}     & \scriptsize{47.6}  & \scriptsize{23.8}    & \scriptsize{4.8}      & \scriptsize{\textbf{19.0}}  & \scriptsize{0}       & \scriptsize{4.8}   \\ \hline
\end{tabular}
\end{table}

\subsection{Human perception of eBear's speech visualization}
In this experiment, we compared the proposed visual speech generation method in Section \ref{mouth_animation} with a na\"{\i}ve lip movement method.  The na\"{\i}ve lip movement method opens and closes the mouth with different weights randomly for each viseme without applying a kernel smoothing and forcing lip closure in labial phonemes (/b/, /m/, /p/). We selected different sentences with various themes including Neutral, Anger, Surprise and Joy. Each sentence lasted 4 to 8 seconds. In one part of the experiment, these sentences were spoken by the robot with na\"{\i}ve lip movements. In the second part, in order to look like actual mouth, lip movements were aligned with the sentences of the speech using the approach explained in Section \ref{mouth_animation}.

12 subjects ages between 21 to 51 years participated in the study. Similar to the previous experiment subjects were selected randomly from various races and cultural backgrounds and were never exposed to the eBear. Each time one speech was visualized twice with two different lip movements starting from neutral face and they were asked to rate how realistic the visual speech looked on a scale from 0 to 5, 0 being unrealistic and 5 being very realistic. By realistic we mean how much the visual speech looks similar to human being speaking, smooth and synchronized with the voice. The speeches were repeated as many times as subjects wished. Tab. \ref{speech_results} shows the experimental results. One-tail paired T-test analysis were conducted and p-values were reported. The results show that the proposed method is significantly better ratted than the na\"{\i}ve method by users.
\begin{table}[htpb]
\caption{User Rating Mean(STD) of Speech Visualization in 0-5 Scale}
\vspace{-0.3cm}
\label{speech_results}
\centering
\begin{tabular}{|c|c|c|c|c|}
\hline
           & \scriptsize{Neutral}      & \scriptsize{Anger}        & \scriptsize{Surprise}     & \scriptsize{Joy}         \\ \hline
\scriptsize{Na\"{\i}ve lip movement }     & \scriptsize{1.58 (0.79)}  & \scriptsize{2.16 (1.64)}  & \scriptsize{1.75 (1.05)}  & \scriptsize{1.41 (0.66)} \\ \hline
\scriptsize{Our method} & \scriptsize{3.75 (0.86)}  & \scriptsize{3.50 (0.67)}  & \scriptsize{3.50 (1.00)}  & \scriptsize{3.91 (0.66)} \\ \hline \hline
\scriptsize{T-test (p-value)}     & \scriptsize{7.83e-6}      & \scriptsize{7.74e-3}      & \scriptsize{6.47e-5}      & \scriptsize{2.88e-8}     \\ \hline
\end{tabular}
\end{table}

\subsection{Facial expression recognition by the eBear vision system}
In the CK+ database, 309 image sequences from 106 subjects are labeled as one of the six basic emotions. In the MMI database, we obtained 209 sessions from 30 subjects. For each of these image sequences, the first frame (Neutral face) and the last three frames (peak frames) were used for expression recognition. The X--Y coordinates of 68 landmark points are given for every image in the CK+ database. For the MMI database, where landmarks are not located, we apply the recent proposed IntraFace \cite{intraface} to detect facial geometry information of images. The X--Y coordinates of the located landmark points were used for image registration via similarity transformation. We cropped the face region from each registered image based on the boundary described by its landmark points, and resized them to $128 \times 128$ pixels. Then, Histogram of oriented gradient (HOG) \cite{hog} and local binary pattern histogram (LBPH) \cite{ahonen2004face} features with $8 \times 8$ windows (no overlap between windows) and $59$ bins in each window were separately extracted from each cropped facial images. Further, for each feature category (LBPH and HOG) the PCA algorithm was used for data dimensionality reduction to preserve 95\% of the energy. RBF and polynomial kernels were utilized for fusion of LBPH and HOG features in our proposed MKL method. 10-fold cross-validation and person-independent schemes were used to evaluate the classification performance. Tab. \ref{ckplus_result} and Tab. \ref{mmi_result} show the confusion matrices of the recognition results on the CK+ and MMI databases, respectively. Compared with several state-of-the-art methods, 92.7\% in \cite{roy2012facial} and 89.3\% in \cite{rivera2012local} on the CK+ database and 86.9\% in \cite{shan2009facial} on the MMI database, our method achieves favorable overall recognition rate (91.2\% on CK+, 89.8\% on MMI).

\begin{table}[htpb]
\caption{Confusion matrix of HessianMKL-based multiclass-SVM with multiple kernels and features on the CK+ database (overall recognition rate: 91.2\%)}
\label{ckplus_result}
\vspace{-0.3cm}
\centering
\begin{tabular}{|c|c|c|c|c|c|c|c|}
\hline
\scriptsize{\%} & \scriptsize{Ag}&\scriptsize{Sp}&\scriptsize{Dg}&\scriptsize{Fr}&\scriptsize{Jy}&\scriptsize{Sd}&\scriptsize{Nt} \\
\hline
\scriptsize{Anger} & \scriptsize{\textbf{91.1}} & \scriptsize{1.5} & \scriptsize{3.7} & \scriptsize{1.5} & \scriptsize{0} & \scriptsize{2.2} & \scriptsize{0} \\
\hline
\scriptsize{Surprise} & \scriptsize{1.2} & \scriptsize{\textbf{94.4}} & \scriptsize{2.4} & \scriptsize{0.8} & \scriptsize{0.8} & \scriptsize{0.4} & \scriptsize{0} \\
\hline
\scriptsize{Disgust} & \scriptsize{5.1} & \scriptsize{0} & \scriptsize{\textbf{84.2}} & \scriptsize{4.5} & \scriptsize{0} & \scriptsize{2.8} & \scriptsize{3.4} \\
\hline
\scriptsize{Fear} & \scriptsize{0} & \scriptsize{2.7} & \scriptsize{4.0} & \scriptsize{\textbf{92.0}} & \scriptsize{0} & \scriptsize{0} & \scriptsize{1.3} \\
\hline
\scriptsize{Joy} & \scriptsize{0.5} & \scriptsize{0} & \scriptsize{1.0} & \scriptsize{1.4} & \scriptsize{\textbf{95.2}} & \scriptsize{0.5} & \scriptsize{1.4} \\
\hline
\scriptsize{Sadness} & \scriptsize{2.4} & \scriptsize{1.2} & \scriptsize{2.4} & \scriptsize{4.7} & \scriptsize{0} & \scriptsize{\textbf{89.3}} & \scriptsize{0} \\
\hline
\scriptsize{Neutral} & \scriptsize{1.3} & \scriptsize{0} & \scriptsize{1.0} & \scriptsize{0.7} & \scriptsize{4.2} & \scriptsize{1.9} & \scriptsize{\textbf{90.9}} \\
\hline
\end{tabular}
\end{table}

\begin{table}[htpb]
\caption{Confusion matrix of HessianMKL-based multiclass-SVM with multiple kernels and features on the MMI database (overall recognition rate: 89.8\%)}
\vspace{-0.3cm}
\label{mmi_result}
\centering
\begin{tabular}{|c|c|c|c|c|c|c|c|}
\hline
\scriptsize{\%} & \scriptsize{Ag}&\scriptsize{Sp}&\scriptsize{Dg}&\scriptsize{Fr}&\scriptsize{Jy}&\scriptsize{Sd}&\scriptsize{Nt} \\
\hline
\scriptsize{Anger} & \scriptsize{\textbf{90.9}} & \scriptsize{0} & \scriptsize{4.1} & \scriptsize{2.0} & \scriptsize{0} & \scriptsize{2.0} & \scriptsize{1.0} \\
\hline
\scriptsize{Surprise} & \scriptsize{1.6} & \scriptsize{\textbf{90.2}} & \scriptsize{3.3} & \scriptsize{4.1} & \scriptsize{0} & \scriptsize{0.8} & \scriptsize{0} \\
\hline
\scriptsize{Disgust} & \scriptsize{6.3} & \scriptsize{1.0} & \scriptsize{\textbf{87.5}} & \scriptsize{3.1} & \scriptsize{0} & \scriptsize{2.1} & \scriptsize{0} \\
\hline
\scriptsize{Fear} & \scriptsize{0} & \scriptsize{2.3} & \scriptsize{3.4} & \scriptsize{\textbf{93.1}} & \scriptsize{0} & \scriptsize{1.2} & \scriptsize{0} \\
\hline
\scriptsize{Joy} & \scriptsize{2.4} & \scriptsize{0.8} & \scriptsize{0} & \scriptsize{2.4} & \scriptsize{\textbf{88.9}} & \scriptsize{0} & \scriptsize{4.7} \\
\hline
\scriptsize{Sadness} & \scriptsize{1.0} & \scriptsize{3.2} & \scriptsize{1.0} & \scriptsize{4.2} & \scriptsize{1.0} & \scriptsize{\textbf{89.6}} & \scriptsize{0} \\
\hline
\scriptsize{Neutral} & \scriptsize{2.4} & \scriptsize{0} & \scriptsize{1.0} & \scriptsize{1.4} & \scriptsize{3.8} & \scriptsize{1.9} & \scriptsize{\textbf{89.5}} \\
\hline
\end{tabular}
\end{table}

To conduct online expression recognition, six novel subjects were asked to express their facial expressions in front of the robot, one minute for each subject. We used IntraFace \cite{intraface} to locate facial landmarks of frame for face registration. The registered image was also saved for manual labeling as the ground truth. HOG and LBPH features were extracted and mapped to lower dimensional spaces via PCA, and then sent to our HessianMKL-based multiclass-SVM classifier for expression labeling. We achieved 77.6\% overall recognition rate when training on CK+ and MMI and testing on the novel subjects. We further trained part of the saved facial images in this online experiment, and tested the rest of the images to conduct person-dependent evaluation, and the overall recognition rate was increased to 98.9\%. On average, in Matlab platform the processes of face detection, feature extraction, and expression classification takes 11 ms per frame using a PC with Intel i5 3.40 GHz CPU. The recognized expression and its number of votes obtained via one-against-one rule are sent to the robot controller application via UDP messages. 

\subsection{Facial expression imitation}
Emotional mirroring is an affect response with one individual imitating other's facial expressions to foster empathy and reinforce the relationship between individuals. We design our robot to automatically imitate user's expressions according to the recognition results from the perception system.

\begin{figure}[h]
\centering
\subfloat[]
{
	\centering
    \includegraphics[scale = 0.14]{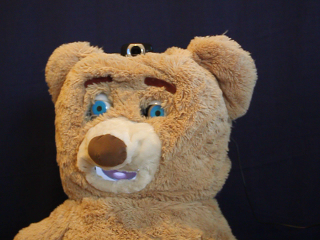}
}
\subfloat[]
{
    \includegraphics[scale = 0.0925]{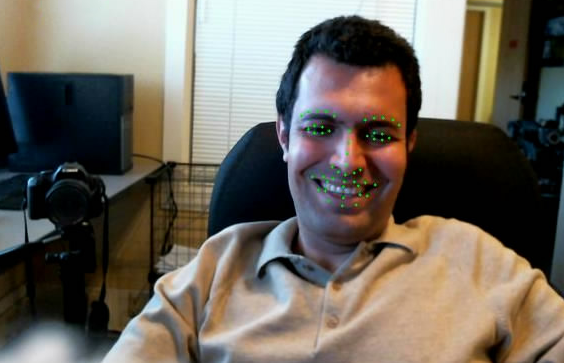}
}
\subfloat[]
{
	\centering
    \includegraphics[scale = 0.14]{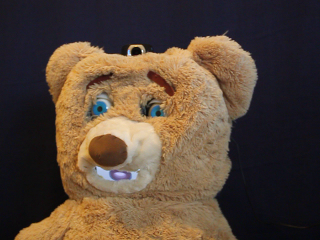}
}
\subfloat[]
{
    \includegraphics[scale = 0.0925]{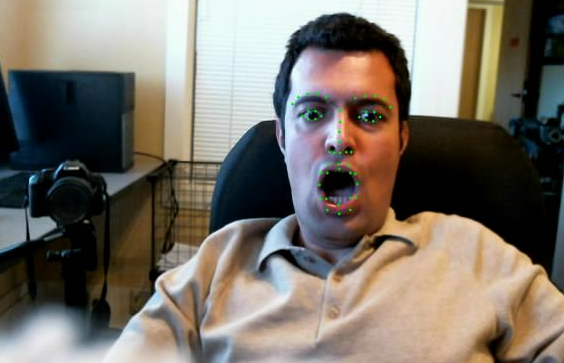}
}
\caption{\label{FacialExpressionImitation}
Examples of facial expression imitation; Joy:(a)(b), Surprise: (c)(d).
}
\end{figure}

One-against-one rule with max-wins voting strategy was utilized in our proposed method for recognizing $P$ human facial expressions. Assuming $Ve_{j}$ indicates the number of votes of the ``winner" (expression $e_{j}$) for a given facial frame, and thereafter $\frac{P-1}{2}\leq Ve_{j} \leq P$. The mapping function that relates the output of robot's perception system to the expression generation system can be formulated as:
\begin{equation}
\label{mapping}
\mu_{i,j} = \lambda_{j} =\frac{2Ve_{j} -P+1}{P-1}, \forall f_{i} \in e_{j}
\end{equation}
where $\mu_{i,j}$ and $\lambda_{j}$ are the intensities of $f_{i}$ in $e_{j}$ in Eqs. \ref{mouth} and \ref{head}, respectively. Based on Eq. \ref{mapping}, the robot is able to control the face DOFs and lips movement for imitating perceived user's expressions, and the intensities of robot's expressions are associated with the credibility of the recognition results. Overall, the performance of the robot in mirroring expressions is limited by the execution time (11 ms) and the online expression recognition accuracy. Fig.~\ref{FacialExpressionImitation} shows some examples of facial expression imitation on the robot.

\section{Conclusion}
\label{concld}
This paper presented the development of a new expressive bear-like robot ``eBear". The eBear face design is hybrid composed of mechanical facial elements with 10 DOFs and a graphical mouth animated on an LCD display. The face can show animal-like facial expressions as well as accurate visual speech for spoken dialog and face-to-face communication with users. User acceptance investigations were utilized to evaluate eBear's expressibility. A novel intelligent framework for recognizing and imitating user's facial expressions were also implemented in our robot platform. Extensive experiments on two public face databases confirm the superiority of our proposed facial expression recognition algorithm compared to several state-of-the-art methods. Overall, our robotic platform enables intuitive human-robot interaction in a mutual setting for real-world applications.

\bibliographystyle{IEEEtran}

\end{document}